\documentclass[runningheads]{llncs}
\usepackage{graphicx}
\usepackage{subfigure}
\usepackage{amsmath,amssymb,amsfonts}
\usepackage{algorithmic}
\usepackage{dirtytalk}
\usepackage{svg}
\usepackage{amssymb}
\usepackage[hidelinks]{hyperref}

\begin{document}
\title{Cellular Plasticity Model for Bottom-Up Robotic Design\thanks{This study was supported in part by the National Science Foundation Graduate Research Fellowship Award \#2136524 to Trevor R. Smith, the National Science Foundation EFRI BRAID Award \#2223793 to Nicholas S. Szczecinski, and the DOD Restoring Warfighters with Neuromusculoskeletal Injuries Research Award (RESTORE)  W81XWH-21-1-0138 to Sergiy Yakovenko. Trevor R. Smith and Thomas J. Smith contributed equally to this work}}

\author{Trevor R. Smith\inst{1} \and
Thomas J. Smith\inst{2} \and
Nicholas S. Szczecinski\inst{1} \and
Sergiy Yakovenko\inst{1} \and
Yu Gu\inst{1}}
\authorrunning{T. Smith et al.}

\institute{West Virginia University, Morgantown, WV 26505, USA \and The University of Texas at Dallas, Richardson, TX, 75080, USA}

\maketitle              

\begin{abstract}

Traditional top-down robotic design often lacks the adaptability needed to handle real-world complexities, prompting the need for more flexible approaches. Therefore, this study introduces a novel cellular plasticity model tailored for bottom-up robotic design. The proposed model utilizes an activator-inhibitor reaction, a common foundation of Turing patterns, which are fundamental in morphogenesis—the emergence of form from simple interactions. Turing patterns describe how diffusion and interactions between two chemical substances—an activator and an inhibitor—can lead to complex patterns and structures, such as the formation of limbs and feathers. Our study extends this concept by modeling cellular plasticity as an activator-inhibitor reaction augmented with environmental stimuli, encapsulating the core phenomena observed across various cell types: stem cells, neurons, and muscle cells. In addition to demonstrating self-regulation and self-containment, this approach ensures that a robot's form and function are direct emergent responses to its environment without a comprehensive environmental model.

In the proposed model, a \textit{factory} acts as the activator, producing a \textit{product} that serves as the inhibitor, which is then influenced by environmental stimuli through consumption. These components are then regulated by cellular plasticity phenomena as feedback loops. We calculate the equilibrium points of the model and the stability criterion. Furthermore, simulations are utilized to examine how varying parameters affect the system's transient behavior and the impact of competing functions on its overall functional capacity. Results show the model converges to a single stable equilibrium tuned to the environmental stimulation. Such dynamic behavior underscores the model's utility for generating predictable responses within robotics and biological systems, showcasing its potential for navigating the complexities of adaptive systems.

\end{abstract}
\section{Introduction and Background}
\label{intro}

Robots today are typically designed through a top-down process to determine their form and function to meet specific design requirements. However, the vast space of potential solutions presents significant challenges in identifying acceptable, let alone optimal, solutions without significantly constraining design freedoms \cite{kang2010approach,nardi2019practical}. Despite extensive research on robotic design, human experience and creativity play vital roles, rendering today's design process arguably more of an art than a science \cite{gericke2011comparisons,tomiyama2009design,kapurch2010nasa}. Because of this synthetic approach, robots often struggle to adapt to the complexities of the natural world \cite{pollack2000evolutionary}. In contrast, biological organisms, which extensively inhabit this world, are shaped by bottom-up interactions within nature, a phenomenon that is challenging to emulate. Nevertheless, the emergence of biological organisms through bottom-up interactions has been extensively studied through the lens of morphogenesis, where low-level chemical interactions between cells lead to complex emergent structures  \cite{mamei2004experiments,sayama2010robust,jin2010morphogenetic}.

A common methodology within morphogenesis research utilizes Turing patterns, which model the diffusion and reaction of two chemicals (activator and inhibitor) as a set of partial differential equations \cite{turing1990chemical}. Subsequent studies have applied this methodology to describe various biological phenomena, such as the formation of limbs, fingers, patterns on seashells, feathers, etc. \cite{marcon2012turing,nakamasu2009interactions,boettiger2009neural}. This concept has been extended to robotics, with systems like Kilobots and Loopy employing Turing patterns for designing their formation from the bottom-up \cite{slavkov2018morphogenesis,smith2023swarm}. For instance, Loopy,  consisting of a closed-loop chain of motor cells, utilizes the activator quantity in each cell to dictate the corresponding motor angle, collectively forming its shape.  These formations demonstrate inherent resilience, such as Loopy correcting self-intersecting shapes and the Kilobots' ability to reform post-damage. While robotic studies have primarily focused on the developmental phase of morphogenesis to generate steady-state robot morphologies, it is essential to recognize that organisms are not static; they constantly adapt their form and behavior to changes in environmental stimuli.

At the cellular level, this adaptability has been captured by cellular plasticity, another aspect of morphogenesis. Cellular plasticity describes how cells enhance their abilities in response to environmental changes.  This phenomenon has been observed throughout many cell types including stem cells \cite{huang2020decoding,kaitsuka2021response,nair2013phylogenetic}, muscle cells \cite{aguilar2019mechanical,feriche2017resistance,de2017role}, and neurons \cite{turrigiano2004homeostatic,la2020brain,jiang2023advances}, each demonstrating unique responses. However, the complexity and high variability of these responses has hindered the development of a simplified, widely applicable model, complicating cellular plasticity's integration into bottom-up robotic design frameworks despite complex models existing for specific biological processes \cite{shen2020cell,oliveri2022mathematical,stiehl2011characterization}.

Recognizing this challenge, our work does not aim to develop a universal model that applies to all facets of cellular plasticity. Instead, we focus on distilling core phenomena from stem cells, muscle cells, and neurons into a simplified model tailored for robotic design. This model aims to replicate key aspects of cellular plasticity: 1) the enhancement of total capacity through specialization - demonstrated by stem cell differentiation  \cite{huang2020decoding,kaitsuka2021response,nair2013phylogenetic}, 2) growth is spurred by product scarcity - demonstrated by muscle cell hypertrophy \cite{aguilar2019mechanical,feriche2017resistance,de2017role}, and 3) exposure to sustained stimuli modulates functional capacity - demonstrated by long-term potentiation/depression of synaptic strength between neurons \cite{turrigiano2004homeostatic,la2020brain,jiang2023advances}. Furthermore, these phenomena, which are governed by self-regulated processes that respond to immediate environmental stimuli, contrast with traditional engineering approaches that depend on fixed setpoints and comprehensive environmental models.

To model the targeted cellular plasticity phenomenon,  an activator-inhibitor reaction, typical of Turing patterns, is augmented with environmental stimuli, thus enabling a uniform description of developmental and cellular plasticity morphogenesis for bottom-up design. By incorporating cellular plasticity into the morphogenesis processes demonstrated on robotic platforms, we could potentially enhance robots' abilities to adapt their emergent form and behavior based on environmental stimuli.  For instance, if the Kilobot robots repeatedly experience damage (i.e., losing neighboring connections), they could increase their density and rate of cohesion. Moreover, to limit the scope of this problem, we will focus on the cellular plasticity of an individual cell. Therefore, our contributions to the literature are:

\begin{itemize}
\item Creation of a simplified cellular plasticity model for the bottom-up design of robots.
\item Expansion of the model to describe multi-functional cells by incorporating various competing processes.
\item Providing analytical and simulation analysis of this model's equilibrium points, stability, and parametric effects on its transient response.
\end{itemize}


\section{Methodology}
\label{methodology}

The proposed cellular plasticity model for bottom-up robotic design includes a \textit{factory}, analogous to an organelle, that produces a \textit{product} molecule consumed by the environment. This factory-product pair represents a cell's ability to perform a function, where the factory's size/quantity dictates the function's intensity, and the product quantity describes the immediate readiness to perform the function. This relationship can be mapped to robotic systems; for instance, a factory-product pair could represent a motor applying torque to a joint in response to angular displacement, with the factory as the torque gain, the product as the available torque, and the consumption rate as the applied torque. Furthermore, the interactions of these components are regulated by the biological phenomena from Section \ref{intro}  manifested as feedback loops. Moreover, to simulate multi-functional cells (i.e., robots with more than one ability), the model is expanded to include multiple pairs of factories and products.

\subsection{Single Factory}

\begin{figure}
    \centering
    \includegraphics[width=.8\textwidth]{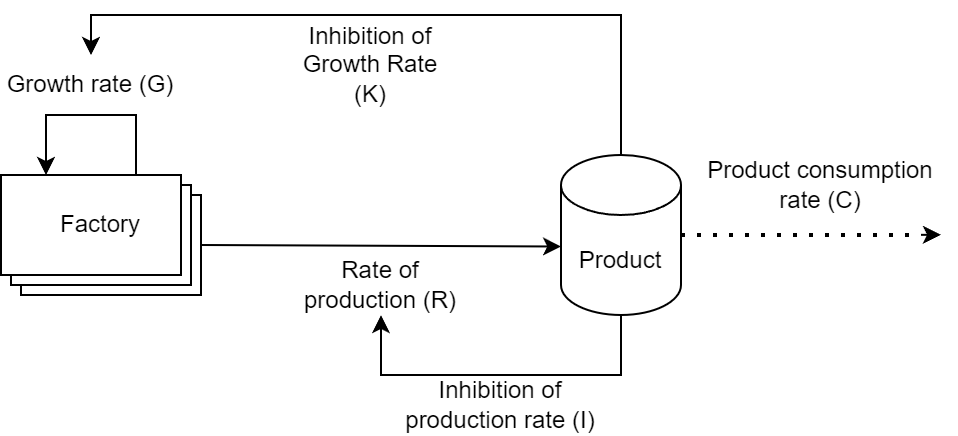}
    \caption{Cellular plasticity model for bottom-up robotic design. This model utilizes a self-replicating factory that produces a product consumed by the environment. This product also inhibits the production and growth rate of the factory, allowing for stable self-regulated adaption to the environmental changes in product consumption.}
    \label{adaptive_model}
\end{figure}

Referring to Fig.~\ref{adaptive_model}, at the heart of the cellular plasticity model is the factory quantity ($F$), which functions as the activator and not only self-replicates at rate $G$ but also produces the inhibitory product quantity ($P$) at rate $R$. This inhibitor modulates the factory by slowing product synthesis and factory growth at respective rates $I$ and $K$. Additionally, the product is consumed by the environment at a rate of $C \cdot P$, where $C$ is the environmental stimulus and the product’s quantity, or availability, proportionally affects how fast it can be consumed. This consumption represents a negative stimulus on the product, which reduces its inhibitory effect on the factory and modulates the equilibrium between the factory ($F$) and product ($P$) quantities with the environmental demand ($C$). This relationship for the cellular plasticity model is mathematically described in (\ref{f_dot}) and (\ref{p_dot}).

 \begin{equation}
    \frac{dF}{dt} = (G - K \cdot P) \cdot F
    \label{f_dot}
\end{equation}

\begin{equation}
    \frac{dP}{dt} = (R - I \cdot P) \cdot F - C \cdot P
    \label{p_dot}
\end{equation}

This model, (\ref{f_dot}) and (\ref{p_dot}), describes two of the three core phenomena of cellular plasticity from Section 1—growth is spurred by product scarcity, and sustained stimuli modulate functional capacity—via two negative feedback loops depicted in Fig. \ref{adaptive_model}. These loops increase the net production rate ($R-I \cdot P$) and net factory growth ($G-K \cdot P$) in response to reduced product levels ($P$), capturing the first phenomenon. Next, if the net factory growth feedback loop reacts slower than the net production rate feedback loop, only prolonged consumption rate changes will significantly modify factory capacity, capturing the second phenomenon. Furthermore, this model captures self-containment and self-regulation, as it does not rely on an external environment model to regulate factory capacity.

\begin{figure}
    \centering
    \includegraphics[width=0.50\textwidth]{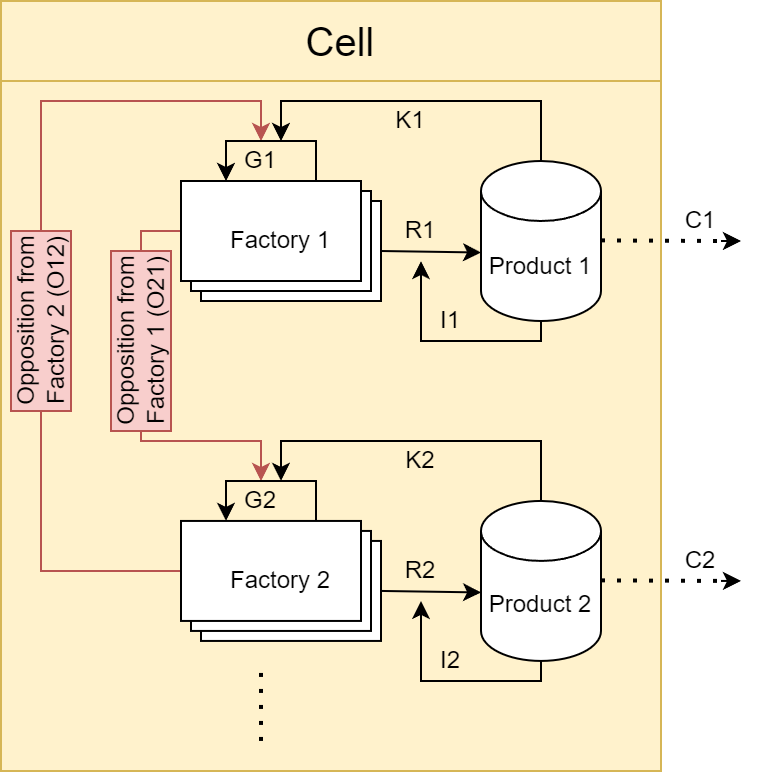}
    \caption{Expansion of the cellular plasticity model to multiple unique functions within a single cell, via multiple competing factories that oppose each other at a rate of $O_{ij}$}
    \label{multi_factory_mech}
\end{figure}

\subsection{Multi-factory}
 
Next, to capture the phenomenon from stem cells that specialization improves total capacity, the cellular plasticity model is expanded to multiple unique processes competing for space in the cell, as shown in Fig. \ref{multi_factory_mech}. This phenomenon is achieved by introducing an opposition rate, $O_{ij}$ that reduces the net growth rate of the $i^{th}$ factory proportionally to the quantity of the opposing $j^{th}$ factory; thus, (\ref{f_dot}) becomes (\ref{multi_f_dot}), where $N$ is the total number of factories in the cell. This opposition rate emulates resource competition, encouraging the cell to specialize by allocating production capabilities to higher-demand products.

\begin{equation}
    \frac{dF_i}{dt} = (G_i - K_i \cdot P_i - \sum_{j \neq i}^{N} O_{ij} \cdot F_j) \cdot F_i
    \label{multi_f_dot}
\end{equation}

\subsection{Model Constraints}

Multiple constraints must be specified to ensure the model captures the cellular plasticity phenomena outlined in Section \ref{intro}. The first constraint is that the parameter values of $G$, $K$, $R$, $I$, and $ O_{ij}$ are positive constants. Secondly, to ensure that only sustained stimuli lead to changes in functional capacity, the model's parameters must be constrained such that the factory's net growth response (i.e., its time constant $\tau_f$), is slower than the production rate's response, $\tau_p$, to changes in the consumption rate ($\tau_f > \tau_p$). Furthermore, the factory ($F$) and product ($P$) quantities are also restricted to positive values, as negative values lack physical meaning. Lastly, the consumption rate stimulus ($C$) is limited to positive values, thus restricting the model from considering external production sources. Section \ref{discussion} further describes the relevance of these constraints in relation to the analyses conducted in Section \ref{Results}.

\section{Analysis}
\label{Results}

Utilizing the methodology from Section \ref{methodology}, this study investigates the system's response to immediate and extended environmental demand surges through step changes in consumption rate. Specifically, this analysis focuses on determining steady-state product and factory levels, stability criteria, and the impact of parameter variations on the time response of the system. Moreover, the study also explores the parametric effects of opposition on the capacities of factories within a multi-factory cell configuration.

\subsection{Steady State and Stability Criteria}

\begin{figure}
    \centering
    \includegraphics[width=.50\linewidth]{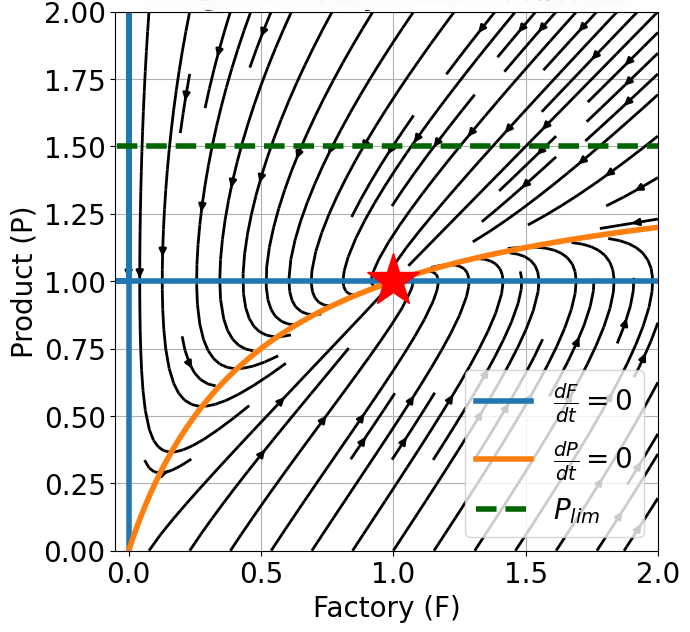}
    \caption{A Representative phase portrait of the single factory's response with the product (orange) and factory (blue) nullclines, and a single non-zero equilibrium (red star) with parameters of $G=K=I=1.0$, $C=0.5$, and $R=1.5$. $P_{lim}$ describes the limit product level as the product nullcline approaches infinity. The inward spiral of the phase portrait signifies a stable equilibrium that attracts from any positive state.}
    \label{phase_portait}
\end{figure}

To get a representative overview of the system dynamics and analyze the effects of initial conditions on the steady-state value of the product and factory, an example phase portrait, with factory and product nullclines (\ref{f_null}) and (\ref{p_null}), respectively, is shown in Fig.\ref{phase_portait}. 

\begin{equation}
   \frac{dF}{dt} = 0 :  P = \frac{G}{K} ,\ \ F = 0
    \label{f_null}
\end{equation}

\begin{equation}
    \frac{dP}{dt} = 0 : F = \frac{C \cdot P}{R-I \cdot P} 
    \label{p_null}
\end{equation}

Analyzing the vector field along the factory and product axes reveals that any state starting in the positive quadrant stays there, ensuring positive factory and product levels. This behavior results from the vertical factory nullcline blocking trajectories from crossing into the negative factory domain and a positive production rate when the product is zero (\ref{p_dot}). Next, by observing the equilibrium point designated by the red star, it is a stable spiral node that attracts all positive states. In addition, due to the horizontal factory nullcline (\ref{f_null}) and inverse proportional product nullcline (\ref{p_null}), the system will only exhibit one non-zero equilibrium point independent of system parameters. Furthermore, by evaluating the denominator of (\ref{p_null}) to zero, we find that the limit of the product nullcline as the factory capacity approaches infinity is finite and equal to $ P_{lim} = R/I$, which is the ratio of the product synthesis rate $R$ and product inhibition rate $I$ and is displayed in Fig. \ref{phase_portait}. 

The steady-state values for product and factory quantities were found by intersecting the nullclines (\ref{f_null}) and (\ref{p_null}), solving for the non-zero equilibrium point (\ref{p_inf}) and (\ref{f_inf}), where the minimum factory quantity required to support the consumption stimulus is $F_{min} = C/R$, which is the ratio of the consumption stimulus $C$ and product synthesis rate $R$.

\begin{equation}
    P_{\infty} = \frac{G}{K}
    \label{p_inf}
\end{equation}

\begin{equation}
    F_{\infty} = \frac{C/R}{K/G - I/R} = \frac{F_{min}}{ 1/P_\infty - 1/P_{lim}}
    \label{f_inf}
\end{equation}

Inspection of (\ref{p_inf}) and (\ref{f_inf}) shows that only the steady-state factory quantity depends on the environmental consumption rate, while the steady-state product level depends on the ratio of intrinsic parameters, specifically the factory growth rate ($G$) and the inhibition of factory growth ($K$). This analysis was extended to the multi-factory steady state using (\ref{p_dot}) and (\ref{multi_f_dot}) to derive (\ref{multi_f_inf}) and (\ref{multi_p_inf}).

\begin{equation}
    F_{i,\infty} = \frac{F_{min,i}}{K_i/(G_i - \sum_{j \neq i}^{N} O_{ij} \cdot F_{j,\infty}) - 1/P_{lim,i}}
    \label{multi_f_inf}
\end{equation}

\begin{equation}
    P_{i,\infty} = \frac{G_i - \sum_{j \neq i}^{N} O_{ij} \cdot F_{j,\infty}}{K_i}
    \label{multi_p_inf}
\end{equation}

In a multi-factory cell, the steady-state levels of the product and the factories are interdependent on the steady-state values of other factories. However, the presence of additional factories invariably reduces the capacity of each factory and product compared to the cell's unopposed single-factory configuration.

Next, to assess the stability of the single factory model, we evaluated the system's Jacobian matrix's trace ($\delta$) and determinant ($\Delta$) at equilibrium points: the origin $(0,0)$ in (\ref{trace_jacob_00}) and the non-zero equilibrium $(F_\infty, P_\infty)$ (\ref{trace_jacob_eq}), respectively. The equilibrium point is stable only if both the trace is negative and the determinate is positive.

\begin{equation}
    \delta_{(0,0)} = G - C < 0 ,\ \ \ \Delta_{(0,0)} = -C \cdot G \ngtr 0 
    \label{trace_jacob_00}
\end{equation}


\begin{equation}
    \delta_{(F_\infty, P_\infty)} = \frac{-C \cdot R/I}{R/I -  G/K} < 0 ,\ \ \ \Delta_{(F_\infty, P_\infty)} = C \cdot G > 0
    \label{trace_jacob_eq}
\end{equation}


Analysis of (\ref{trace_jacob_00}) shows the origin is inherently unstable due to the always negative determinate, thus directing the system towards positive values. However, for the non-zero equilibrium $(F_\infty, P_\infty)$ to be stable (\ref{trace_jacob_eq}) must have a positive denominator, establishing the stability criterion (\ref{stab}). Where the steady state product level ($P_\infty$) must be less than the limit product level ($P_{lim}$)

\begin{equation}
    \frac{G}{K} = P_\infty < \frac{R}{I} = P_{lim}
    \label{stab}
\end{equation}

\subsection{Parametric Effects on Time Constants}
\begin{figure}
    \centering
    \includegraphics[width=.55\linewidth]{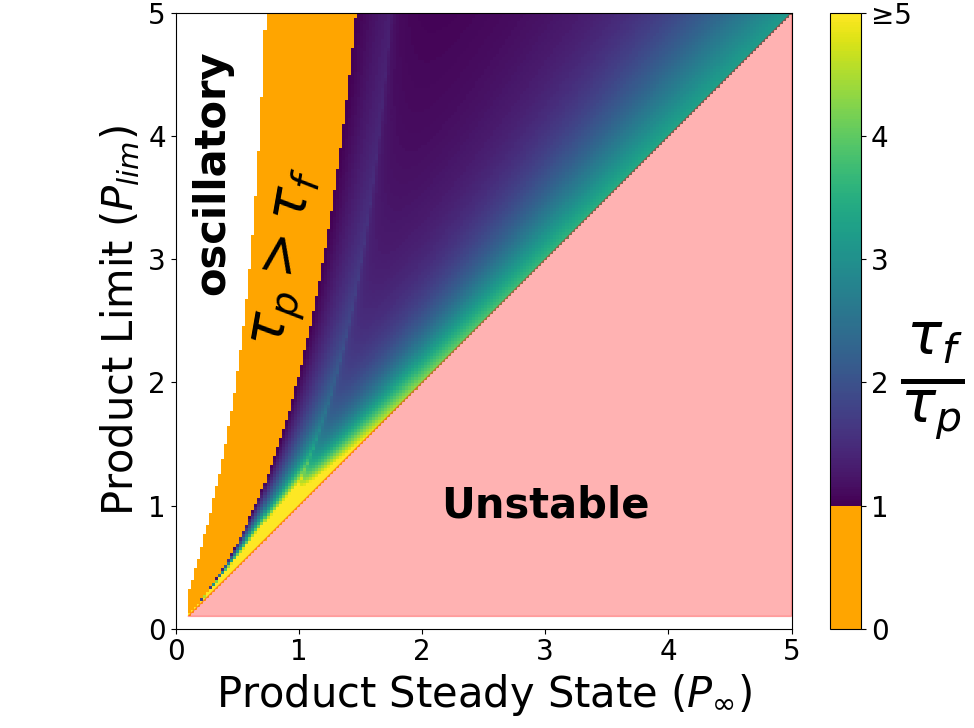}
    \caption{Parametric effects on the relative time constants of the product ($\tau_p$) and factory ($\tau_f$), particularly as system parameters near zero or instability $\tau_f$ becomes much larger than $\tau_p$. The orange region displays where the product time constant exceeds the factory time constant. Damped oscillatory modes emerge when $P_{lim}$ significantly exceeds $P_\infty$, marked by the white region.}
    \label{fig_time_const}
\end{figure}

Now that the stability criterion and steady-state values for the factory and product have been established, the parametric effects on the model dynamics are examined via simulations to evaluate the time constants of the factory ($\tau_f$) and product ($\tau_p$) in response to step changes in consumption rate. To simplify analysis The system is reparameterized from $(G, K, R, I, C)$ to steady-state product level ($P_{\infty} = G/K$), limit product level ($P_{lim} = R/I$), and minimum factory quantity ($F_{min} = C/R$), by dividing (\ref{f_dot}) by $G$ and (\ref{p_dot}) by $R$. This parameter reduction simplifies analysis from five parameters to two when applying a unit step input to $F_{min}$. In Fig. \ref{fig_time_const},  $P_{lim}$ and $P_{\infty}$ were varied from [0.1 to 5] with the time constant ratio ($\tau_f / \tau_p$) plotted as a heatmap.

From Fig.\ref{fig_time_const}, it is observed that as $P_{lim}$ and $P_\infty$ approach zero and the instability threshold, the relative time constant between the factory and the product becomes very large. As $P_{lim}$ and $P_\infty$ move away from instability, the factory time constant eventually becomes smaller than the product time constant (orange region in Fig.\ref{fig_time_const}). This indicates that the factory responds faster than the product, adapting to short-term changes while the product adapts to long-term changes, violating the core phenomenon of sustained stimuli modulating functional capacity. Furthermore, as $P_{lim}$ increases relative to $P_\infty$, the system exhibits a damped oscillatory response (white region in Fig.\ref{fig_time_const}). This causes the factory capacity ($F$) to overshoot its steady state, leading to an over-response to stimuli. Despite being stable when $P_\infty << P_{lim}$, the model constraint that the factory time constant be slower than the product time constant ($\tau_f > \tau_p$) further restricts the valid parameter range to the heat map region.

\subsection{Transient Response to Short- and Long-Term Changes in Consumption}

\begin{figure}
    \centering
    \includegraphics[width=.56\textwidth]{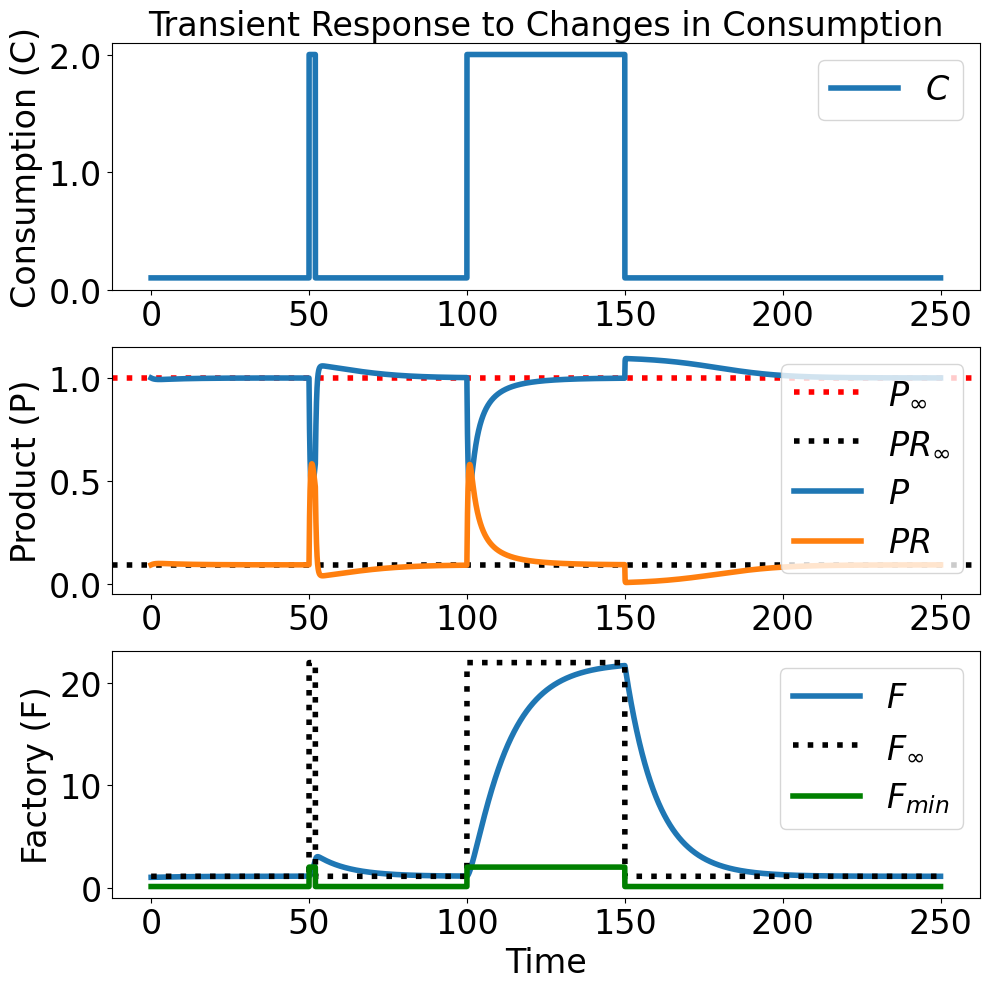}
    \caption{Product ($P$) and Factory ($F$) transient response to short (at $t=50$) and long (at $t=100$) periods of increased in consumption (C). In addition to the net production rate ($PR$) and minimum factory capacity ($F_{min}$). Furthermore, the steady-state factory, product, and production rate are plotted with dotted lines. Parameter values are $G=K=I=1.0$ and $R=1.1$}
    \label{fig_trans}
\end{figure}

Next, to analyze the system response to short- and long-term changes in the consumption rate of the environment, the system with parameter values of $G = K = I = 1.0$ and $R = 1.1$ was subjected to a significant increase in the consumption rate (2 product/time unit) for a short duration (2 time units) and long duration (50 time units) and plotted in Fig. \ref{fig_trans} along with the net production rate $PR = R-I \cdot P$, the minimum factory quantity $F_{min}$, and steady state factory ($F_\infty$) and product levels ($P_\infty$).

The first observation of Fig. \ref{fig_trans} is that the model exhibited distinct behaviors on different time scales of resource fluctuation. For short-term deficits (at $t=50$),  the production rate increased rapidly with only a small increase in factory capacity. However, if the deficit is prolonged (at $t=100$), only then will the factory capacity increase significantly. Furthermore, a higher quantity of the factory persists after the removal of the heightened consumption rate until it eventually decays and the factory capacity returns to the new steady state of the environment. The following observation of Fig. \ref{fig_trans} shows that the steady-state factory level is approximately ten times larger than the minimum required to support the consumption rate.

\subsection{Effects of Opposition on Identical Factories}

\begin{figure}
    \centering
    \includegraphics[width=1\linewidth]{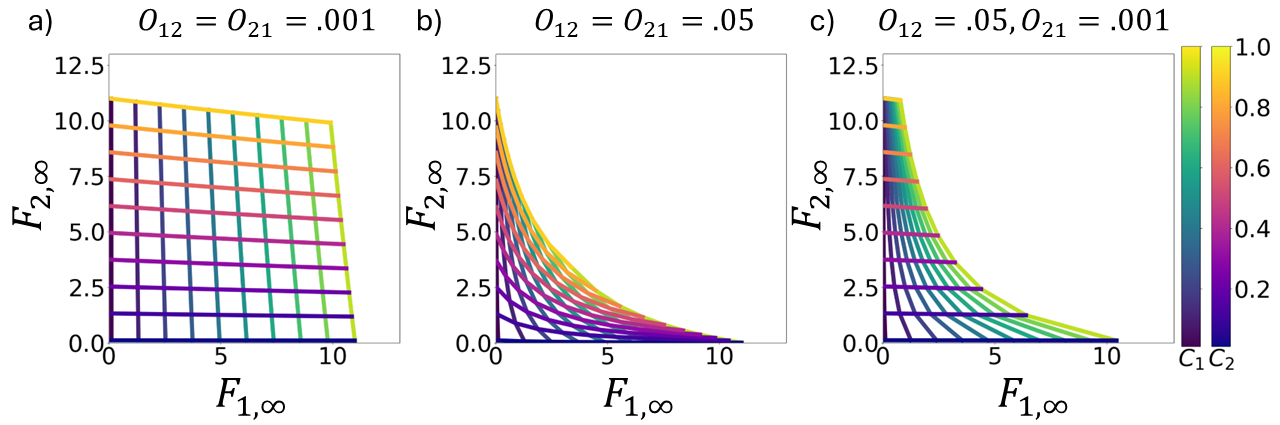}
    \caption{Effects of low (a), high (b), and asymmetrical (c) opposition rates ($O_{ij}$) on the steady-state factory quantities of a multi-factory cell. More vertical lines are curves of constant consumption of factory 1's product ($C_1$), while more horizontal lines are curves of constant consumption of factory 2's product ($C_2$)}
    \label{fig_opp_eqlib}
\end{figure}

To assess the opposition parameter's impact on factory steady-state levels, we analyzed two factories with identical parameters under varying consumption rates ([.1 - 1]) across three scenarios: low-symmetric opposition ($O_{12} = O_{21} = .001$), high-symmetric opposition ($O_{12} = O_{21} = 0.05$), and asymmetric opposition ($O_{12} = 0.05, O_{21} = .001$). Fig. \ref{fig_opp_eqlib} displays the factory steady-state levels for each of these scenarios at the intersection of the curves of constant consumption of each product: vertical curves for $C_1$ and horizontal curves for $C_2$.

The initial analysis of Fig. \ref{fig_opp_eqlib} reveals that with rising symmetric opposition rates, the constant consumption lines transition from nearly square shapes—indicating no inter-factory dependence—to shapes increasingly skewed towards zero. This deformation indicates that high opposition rates significantly restrict factory capacities when both products have high consumption rates; however, the impact is minimal if either factory's product consumption rate is low. Furthermore, applying opposition asymmetrically results in the steady-state capacity surface stretching towards the factory with lesser opposition. This adjustment suggests that asymmetric opposition directly influences factory capacities, favoring the less opposed factory in the allocation of resources.

Next, to analyze the effects of opposition on the total capacity of the multi-factory cell, the consumption rates for a cell with two identical factories were varied from [0.1 to 1.0], and the total capacity was plotted as a heat map along with contours of total consumption ($Totcon = C_1 + C_2$) and total capacity ($Total\ \ capacity = F_1 + F_2$).

\begin{figure}
    \centering
    \includegraphics[width=.695\linewidth]{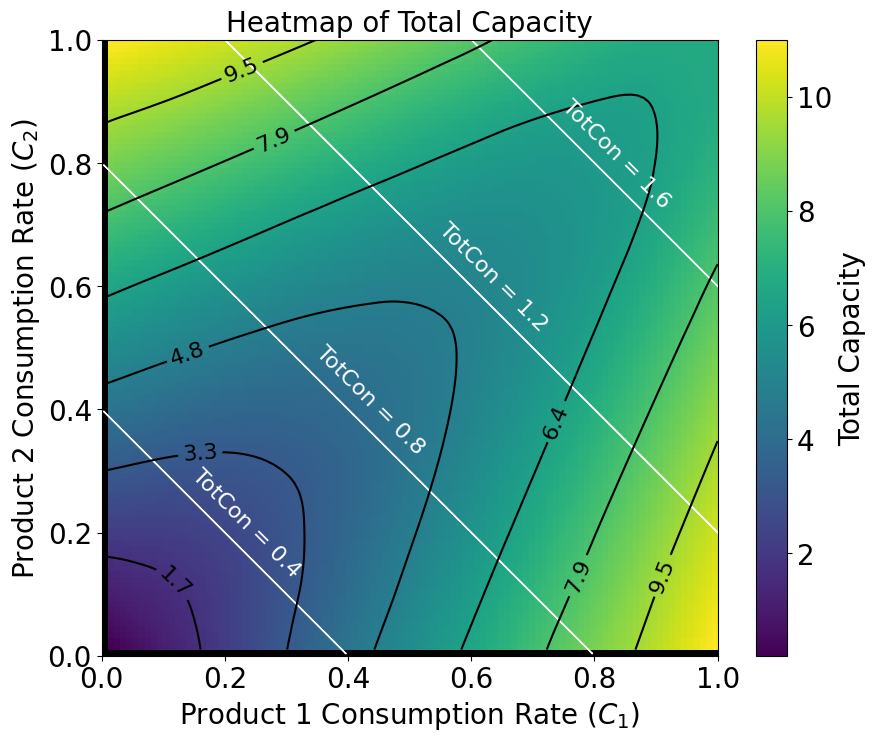}
    \caption{Heatmap of the total capacity of a two factory cell relative to the consumption rates of factory 1 ($C_1$) and factory 2 ($C_2$). Black lines are contour lines of total factory capacity, while white lines are contours of constant total consumption (TotCon)}
    \label{fig_tot_cap}
\end{figure}

In Fig. \ref{fig_tot_cap}, it is evident that as a cell specializes (increases one factory's capacity over the other), its total capacity rises. This observation is readily seen by tracing the total consumption line equal to 0.8 from the center towards either factory 1 or 2, where the total capacity grows from approximately 4 to over 7.9. 

\section{Discussion}
\label{discussion}
Based on the results of Section \ref{Results}, the proposed cellular plasticity model imitates the generalized biological phenomena: 1) growth is spurred from product scarcity, 2) sustained stimuli modulate functional capacity, and  3) specialization increases total capacity, in addition to being self-contained and self-regulating, provided its constraints are met. However, the model's constraint for positive factory and product quantities is mild due to any initially positive factory and product pair remaining positive, as shown in Fig. \ref{phase_portait}. Furthermore, the constraint requiring slower factory growth compared to product synthesis is partially mitigated by Fig. \ref{fig_time_const}, illustrating a range of valid parameter values.

The proposed model displays how growth is driven by product scarcity in (\ref{f_inf}) and (\ref{multi_f_inf}), where the factory's steady-state capacity expands in proportion to the environmental demand ($C$). This behavior is visualized in Fig. \ref{fig_trans}, where factory growth is initiated only when product levels fall below their steady-state value. This response is reminiscent of muscle strengthening, where cells increase oxygen uptake or Adenosine triphosphate (ATP) production in response to deprivation \cite{feriche2017resistance,de2017role}. Furthermore, this approach emphasizes self-containment, where adaptations rely on internal factors (e.g., factory and product quantities) rather than a simplified environment model. By using real-world stimuli to influence product levels and factory capacities, the robot could adapt to the complexities of unpredictable real-world conditions that a human-specified model might neglect.

The phenomenon of functional capacity modulation by sustained stimuli is effectively demonstrated in Fig. \ref{fig_trans}, where only prolonged demand changes influence factory capacity, not short-term spikes. Initially, the system leverages on-hand product reserves to address sudden increases in demand $C$, subtly adjusting its production rate to meet the new consumption rate. However, a significant increase in factory capacity only occurs if the product deficit is prolonged, similar to the biological mechanism of long-term potentiation/depression, where sustained activity is essential for synaptic strength modulation \cite{turrigiano2004homeostatic,jiang2023advances}. This dual-response adaptation, along with the system's ability to exceed minimal needs (Fig. \ref{fig_trans}), equips the cell for prolonged and dynamic demand surges without excessively responding to temporary spikes.

Also, the model exhibits self-regulation by converging to a stable equilibrium specifically tuned to environmental stimulation, as illustrated in Fig. \ref{phase_portait} and supported by equations (\ref{p_inf}) and (\ref{f_inf}). 
This single equilibrium shows that a cell's capacity directly responds to environmental stimuli, reflecting the environment's impact on cellular functions. Such dynamic behavior underscores the model's utility for generating predictable responses within robotics and biological systems, showcasing its potential for understanding these complex systems.

The next phenomenon, specialization increases total capacity, is exhibited by this model due to the single factory steady state capacity (\ref{f_inf}) being higher than the multi-factory steady state capacity (\ref{multi_f_inf}), thus the highly specialized single factory cell has more capacity than a generic multi-factory cell. In addition, this insight is displayed in Fig. \ref{fig_opp_eqlib} by the steady-state factory capacity being largely unaffected by the opposition of the other factory if the consumption rate is much higher for one compared to the other. Furthermore, from Fig. \ref{fig_opp_eqlib}, the asymmetric opposition rates predispose the cell to specialize into the lesser opposed factory. This response is similar to the histone effects on stem cells, where the cells are pre-biased to form into their parent cell types \cite{nair2013phylogenetic}. Furthermore, from Fig. \ref{fig_opp_eqlib}, the deformation of the upper right corner of the equilibrium surface towards zero emulates a limited-resources constraint, where it is more difficult to perform two tasks equally than to specialize in one task. Moreover, Fig. \ref{fig_tot_cap} directly displays that the model captures this phenomenon due to the total capacity of the cell being minimal when the consumption rate is equal between the two factories (i.e., on the 45-degree diagonal). This specialization mechanism underscores the importance of adaptability and efficiency in robotics, indicating that robots initially equipped with multiple functions can dynamically adjust their functional capacity to prioritize higher-demand functions, effectively designing themselves from the bottom up.

Since this model utilizes the activator-inhibitor framework, it can be integrated into existing morphogenesis-based bottom-up design frameworks such as \cite{slavkov2018morphogenesis,smith2023swarm}, potentially similar to the multi-factory cell configuration. A simple application of this model is in adapting the rigidity of a robot's morphology to the environment. Specifically, this could be done utilizing the Loopy platform, which is constructed as a chain of servo motors \cite{smith2023swarm}, by modifying the response of each motor to its angular error. Therefore, in this scenario, the factory could be a proportional gain to the angular error, and the consumption rate could be the supplied torque, with the product representing the available torque. Thus, based on (\ref{f_inf}), the servo motor's response to perturbations would vary with the external load: under lighter loads, it would demonstrate increased flexibility, whereas, under heavier loads, its response would become more rigid due to a higher gain. Furthermore, this example can be extended to the multi-factory case, where each motor also regulates its response to angular velocity, thus modifying Loopy's viscosity (i.e., the fluidity of its response to perturbations). Thus, macroscopically, the rigidity and viscosity of Loopy's morphology would be designed from the bottom up as an emergent response to its environment. This application highlights the model's potential to inspire further bottom-up designs for more responsive and adaptable robotic systems based on principles observed in cellular plasticity.

\section{Conclusions and Future Work}
\label{conclusion}
In this work, we successfully developed a cellular plasticity model for morphogenesis-based bottom-up robotic design. This model captures general phenomena observed in various cell types: 1) growth stems from product scarcity from muscle cells, 2) sustained stimuli modulate functional capacity from neurons, and 3) specialization increases total capacity from stem cells, in addition to being self-contained and self-regulating. Furthermore, the singular stable equilibrium point for a given environment demonstrates that a cell's distribution of functional capabilities is a direct emergent response to the environment. Moreover, by utilizing the activator-inhibitor framework augmented with environmental stimuli, this model can be readily incorporated into existing morphogenesis-based bottom-up robotic designs. Therefore enhancing robots’ abilities to adapt their emergent form and behavior to their environment.

This study is limited by simplifying complex biological processes to general phenomena and assuming that factories can grow indefinitely given adequate time. Additionally, it focuses solely on how a single cell adapts its morphology without analyzing diffusion effects or external production sources of the product. Furthermore, the lack of experimental validation with physical robots restricts the empirical confirmation of the model's applicability. Future endeavors aim to address these gaps by undertaking experimental validations with physical robots to examine the influence of cellular plasticity on the robot's emergent morphology. In addition to enhancing the model to include a defined factory capacity limit per cell and incorporating a model for cellular division and death. Furthermore, this model will adjust the parameters of the spawning cell (similar to evolutionary algorithms \cite{floreano2000evolutionary} and stem cell histones \cite{nair2013phylogenetic}) to bias specialization towards the additional capacity needed by the parent cell, introducing another layer of adaptive response at a longer time scale. Moreover, this work will extend to multiple cells with varied parameters and redundant factories, incorporating diffusion effects to better emulate multicellular behaviors and possibly create Turing patterns. The goal is to explore specialization and collaboration among cells to aid multi-robot teams in task selection and specialization from the bottom up, making them more responsive and versatile.


\bibliographystyle{IEEEtran}
\bibliography{references}

\end{document}